\documentclass[runningheads]{llncs}
\usepackage{graphicx}
\usepackage{amsmath,amssymb} 
\usepackage {algorithmic}
\usepackage[ruled,commentsnumbered]{algorithm2e}
\usepackage{color}
\usepackage{booktabs}
\usepackage[width=122mm,left=12mm,paperwidth=146mm,height=193mm,top=12mm,paperheight=217mm]{geometry}
\usepackage{wrapfig}

\begin{document}

\newcommand{\n}{\vec{\eta}}
\newcommand{\I}{I}
\newcommand{\D}{\Omega}
\newcommand{\R}{\mathbb{R}}
\newcommand{\Patch}{P}
\newcommand{\C}{\mathcal{C}}
\newcommand{\flow}{\nu}
\newcommand{\etal}{et al.}
\newcommand{\todo}[1]{\textcolor{red}{TODO:#1}}
\newcommand*\rot{\rotatebox{90}}

\pagestyle{headings}
\mainmatter

\title{Deep Active Contours} 

\titlerunning{Deep Active Contours}

\authorrunning{Rupprecht, Huaroc, Baust, Navab}

\author{Christian Rupprecht$^{1,2}$, Elizabeth Huaroc Moquillaza$^1$, Maximilian Baust$^1$, Nassir Navab$^{1,2}$}
\institute{$^1$Technische Universit\"at M\"unchen, Munich, Germany\\$^2$Johns Hopkins University, Baltimore, USA}

\maketitle

\begin{abstract}
	We propose a method for interactive boundary extraction which combines a deep, patch-based representation with an active contour framework.
	We train a class-specific convolutional neural network which predicts a vector pointing from the respective point on the evolving contour towards the closest point on the boundary of the object of interest.
	These predictions form a vector field which is then used for evolving the contour by the Sobolev active contour framework proposed by Sundaramoorthi et al.
	The resulting interactive segmentation method is very efficient in terms of required computational resources and can even be trained on comparatively small graphics cards.
	We evaluate the potential of the proposed method on both medical and non-medical challenge data sets, such as the STACOM data set and the PASCAL VOC 2012 data set.
\end{abstract}

\section{Introduction}
\label{sec:Introduction}
The goal of the presented paper is to develop an interactive method for image segmentation based on an active contour framework, where the contour evolution is governed by predictions from a convolutional neural network (CNN).
\subsection{Motivation}
\label{sec:Motivation}
The presented work combines the power of CNNs with high level mechanisms for inference.
Very recent examples for this combination are the works of Zhang \etal \cite{Zhang15,Zhang16}, which combine a patch-based deep-representation with Markov random fields (MRFs) in order to obtain an accurate instance level segmentation of road scenes, or the work of Ranftl and Pock \cite{Ranftl14}, where the unary and pairwise potentials for a global inference model are connected to the output of a CNN.

However, applications of CNNs to image segmentation, such as the recent works of Ronneberger \etal~\cite{Ronneberger15} or Badrinarayanan \etal~\cite{badrinarayanan2015segnet2} for example, typically require the input images to be of fixed size and demand large models with tens to hundreds of millions of parameters. 
By combining a patch-based deep representation with an active contour model, we are able to segment images of arbitrary size and -- even more importantly -- perform the patch extraction process only in a vicinity of the object of interest.
This way, we obtain a very flexible and efficient interactive image segmentation method.
\subsection{Related Work}
\label{sec:RelatedWork}
As the proposed method combines ideas from two worlds, i.e. active contours and deep learning, we split the review of related works according to these categories.
\paragraph{Active Contours}
Since their introduction by Kass, Witkin and Terzopoulos in \cite{Kass88}, Active Contours have become a very popular tool for image segmentation.
In their original form, active contours are however a very local technique in the sense that their evolution is governed by weak and unspecific features, i.e. edge information, extracted in the vicinity of the evolving curve.
Thus, much effort has been made in order to increase the robustness of active contours as well as to extend their applicabilities.
Well-known examples are the works on geodesic active contours by Yezzi \etal~\cite{Yezzi97tmi} and Caselles \etal~\cite{Caselles97}, on gradient vector flow by Xu and Prince \cite{Xu98}, on incorporating region-based information by Zhu and Yuille \cite{Zhu96}, Tsai \etal~\cite{Tsai01} and Chan and Vese~\cite{Chan01}, on incorporating shape information by Cootes and Taylor \cite{Cootes95}, Leventon \etal~\cite{Leventon00}, Tsai \etal~\cite{Tsai03}, or on incorporating different cues such as color, (dynamic) texture or motion by Soatto \etal~\cite{Soatto01}, Rousson \etal~\cite{Rousson03}, and Cremers and Soatto \cite{Cremers05}.
Of course, all efforts on adapting convex relaxation techniques, such as Chan \etal~\cite{Chan06} or Bresson \etal~\cite{Bresson07}, deserve to be mentioned in this context, too. 
Besides using convex relaxation techniques, it is also possible to employ different (regularizing) metrics in the space of curves in order to make the contour evolution more robust: popular examples are Charpiat \etal~\cite{Charpiat07}, Sundaramoorthi \etal~\cite{Sun07}, or Melonakos \etal~\cite{Melonakos08}.
In this work, we employ some ideas of the so-called Sobolev active contours proposed in \cite{Sun07} due to their excellent regularizing properties.

From a conceptual point of view, also the works of Lankton \cite{Lankton08} and Sundaramoorthi \cite{Sun07} can be seen as related, because their approaches for localizing active contours consider the local neighborhood of an evolving contour point.

As this work employs CNNs for steering the contour evolution, it is also important to mention that there exist also very early works which employed neural networks as an optimization framework in order to evolve the contour, see Villarino \cite{Villarino98} or Rekeczky and Chua \cite{Rekeczky99} for example.
The employed networks, however, are neither deep, nor are used for training features of certain object categories.
\paragraph{CNNs for Image Segmentation}
CNNs have been used extensively for image segmentation. 
For example the subtask of semantic scene parsing has been tackled with deep networks. 
Farabet \etal~\cite{farabet2012scene,farabet2013learning} extract a hierarchical tree of segments which are assigned a feature by using a CNN.
The extracted segments are then classified based on this feature exploiting the tree structure. 
Gatta \etal~\cite{gatta2014unrolling} introduce semantic feedback into the CNN prediction to improve pixel-wise classification results. 
More recently, encoder-decoder structures, that first down-sample the input image and then up-sample it have been employed for image segmentation. 
Neh \etal~\cite{noh2015learning} and Badrinarayanan \etal~\cite{badrinarayanan2015segnet2} use the pooling location from the down-sampling state for un-pooling in the up-sampling part of the network to reproduce finer details in the prediction.
Fully convolutional CNNs were first used by Long \etal~\cite{long2015fully} for segmentation tasks who also introduce skip layers to combine high level information with earlier, higher resolution feature maps to sustain finer details in the output segmentation.
Ronneberger \etal~\cite{Ronneberger15} make extensive use of those skip connection to propagate information between all stages during down and up-sampling.
Our approach differs from the aforementioned methods in two ways.
Firstly, basing on active contours the proposed method is interactive in contrast to the cited CNN related methods.
Secondly, as coarse object location is determined by the initialization, we do not need to sample the whole image to evolve the contour providing a much higher scalability of our method with increased image size.
\subsection{Contribution}
\label{sec:Contribution}
This paper proposes an interactive image segmentation technique based on an active contour framework.
Thereby, the contour evolution is governed by a vector field which is predicted by a CNN.

More precisely, we rely on an explicit contour representation and sample small patches at each point on the evolving contour.
These patches are then used to predict a vector pointing from the patch center to the closest object boundary.
This way, we obtain a vector field attached to the curve which is then used to evolve the contour by a Sobolev active contour framework, cf.~\cite{Sun07}.
The advantages of this method are as follows:
\begin{enumerate}
	\item \textbf{Logarithmic Complexity:} As we only sample patches around the evolving contour, our method scales approximately logarithmically with the size of the object, cf. Sec.~\ref{sec:Gradient}.
	\item \textbf{Efficient and Fast Training:} The employed CNN is relatively small in comparison to other networks used for image segmentation.
	In fact we can train the employed CNN on an NVIDIA GTX 980 graphics card with only 4GB of RAM within two hours.
	\item \textbf{Broad Applicability:} The proposed method can be trained for virtually any segmentation application.
	Furthermore, we require relatively few image sets for training the network which is of particular interest for medical applications.
	For an example, the network which has been used for the left ventricle segmentation,  cf. Sec.~\ref{sec:STACOM} could be trained using only 30 short axis MRI scans.
\end{enumerate}
To our best knowledge, this is the first work combining deep convolutional neural networks and active contours for interactive image segmentation.

\section{Methodology}
\label{sec:Methodology}
%
We start by reviewing and generalizing  the classical active contour recipe in Sec.~\ref{sec:Generalized}, before outlining the details on the CNN architecture and training in Sec.~\ref{sec:Gradient} and Sec.~\ref{sec:Architecture}.
\subsection{Generalized Active Contour Framework}
\label{sec:Generalized}
As noted in \cite{Kass88}, active contours are called \textit{active}, because they are minimizing an energy and thus exhibit dynamic behavior.
In this work, we consider a slightly more modern, i.e. geometric, formulation of active contours, see \cite{Caselles97} or \cite{Sun07} for instance. The standard active contour recipe reads:
Let
\begin{equation}
\C:\begin{cases}
[0,L]\times[0,\infty) &\rightarrow\R^2,\\
(s,t)&\mapsto \C(s,t),
\end{cases}
\end{equation}
denote a temporally varying family of closed curves parametrized w.r.t.~arc length $s$.
The total length of the curve is denoted by $L$.
In order to segment an object of interest, one usually defines an energy
\begin{equation}
E(\C)=D(\C,I)+\lambda R(\C),
\end{equation}
where $D$ is a data term, that measures how well the curve $\C$ segments the image $I:\D\rightarrow[0,1]^d$ ($d=1$ for gray scale images and $d=3$ for color images), $R$ is a regularization term, which usually penalizes the length or curvature of $\C$, and $\lambda>0$ is a positive parameter controlling the trade-off between regularization and data fidelity.
The desired configuration of the curve usually corresponds to a local minimum of $E$ which makes an initial placement of the contour by some "higher-level processes" \cite{Kass88}, such as a manual initialization for example, necessary.
Once initialized, the final configuration of the contour is found by advecting the contour according to the gradient flow
\begin{equation}
\partial_t \C(s,t)=\nabla E(\C(s,t)),
\label{eqn:gradFlow}
\end{equation}
where $\nabla E(\C(s,t))$ is a vector field attached to curve.
This vector field is usually found by computing $\left. \partial_t E(\C(s,t))\right|_{t=0}$.
It is important to note the following two points:
\begin{enumerate}
	\item As one is usually only interested in the \textit{geometry} of the final curve, only the normal part of $\nabla E(\C(s,t))$ is used for the evolution.
	Thus a more general form of \eqref{eqn:gradFlow} is
	\begin{equation}
	\partial_t \C(s,t)=\alpha(s,t)\n(s,t),
	\label{eqn:gradFlowGeneralized}
	\end{equation}
	where $\n$ denotes the outer normal of $\C$ and
	\begin{equation}
	\alpha(s,t)=\nabla E(\C(s,t)) \cdot \n
	\label{eqn:speedFunction}
	\end{equation}
	is the so-called speed function.
	\item The notation $\nabla E(\C(s,t))$ suggests that $\nabla E(\C(s,t))$ is \textit{the} gradient of $E$.
	This interpretation is, however, wrong as noted in \cite{Charpiat07,Sun07,Yezzi05}: the gradient depends on the underlying Hilbert space used for computing the $\left. \partial_t E(\C(s,t))\right|_{t=0}$.
	In fact, we will make use of this observation in this paper, too.
\end{enumerate}
In contrast to such a classical approach, however, the active contours in this work are not energy-minimizing.
Instead, we are going to train a CNN which predicts at each point on the discretized contour a vector which points to the boundary of the closest object instance for which the CNN has been trained.
Thus, we term our active contours \textit{deep active contours}.
If for an example, the deep active contour has been trained on airplanes, the predicted vector will point from the contour towards the closest pixel on the boundary of an airplane.
Thus, we predict in fact a vector field $\flow(s,t)$ at time $t$ based on which the speed function in \eqref{eqn:gradFlowGeneralized} is computed: $\alpha(s,t)=\flow(s,t)\cdot \n(s,t)$.
Although $\flow(s,t)$ is not a vector field related to the minimization of an energy, we can still make use of a Sobolev-type regularization, see Sundaramoorthi \etal \cite{Sun07}, and employ
\begin{equation}
\flow(s,t)\ast K_\beta(s),\quad\text{where}\quad K_\beta(s)=\frac1L\left(  1+L\frac{(s/L)^2 - (s/L)+1/6}{2\beta} \right) 
\label{eqn:SobolevKernel} 
\end{equation}
and $\beta=0.01$.
By using this type of regularization, we encourage translations of the contour or smooth deformations causing the resulting contour evolution to be even more robust with respect to spurious local predictions.
The resulting algorithm for evolving the proposed deep active contour is outlined in Algorithm \ref{alg:DeepSnakeEvolution}.
\IncMargin{-0.3em}
\begin{algorithm}[t]
	\SetCommentSty{footnotesize}
	\SetKwFunction{predVector}{predVector}
	\SetKwFunction{samplePatch}{samplePatch}
	\SetKwFunction{converged}{converged}
	\SetKwFunction{regularizeVectorField}{regularizeVectorField}
	\SetKwInOut{Input}{input}
	\SetKwInOut{Output}{output}

	\Input{Image $I$, number of steps $N$, stepsize $\tau$, number of curve points $K$, initialization of contour $C(s,0)=(C^0_1,\ldots,C^0_K)$}
	\Output{final segmentation $\C(s,T)=(C^N_1,\ldots,C^N_K)$ for $T=N\tau$.}
	\BlankLine
	\For{$i\leftarrow 1$ \KwTo $N$ }
	{
		\tcp{predict vector field}
		\For{$j\leftarrow 1$ \KwTo $K$}
		{
		$P_j \leftarrow$  \samplePatch{$C^i_j$,$I$}\tcp*{cf.~Sec.~2.2}
		$\flow^i_i \leftarrow$  \predVector{$P^i_j$}\tcp*{cf.~Sec.~2.3}
		}
		$\flow^i \leftarrow$  \regularizeVectorField{$\flow^i$}\tcp*{cf.~Eq.~(5)}
		\tcp{evolve contour}
		\For{$j\leftarrow 1$ \KwTo $K$}
		{
			$C^{i+1}_j \leftarrow C^{i+1}_j + \tau  \flow^i_j\cdot \n^i_j $
		}
		\If{\converged{$C^{i+1}, C^i$}}{break}
	}
	\caption{\textbf{Deep Snake Evolution}}
	\label{alg:DeepSnakeEvolution}
\end{algorithm}
\subsection{Predicting the Flow}
\label{sec:Gradient}
\begin{figure*}[t]
	\begin{center}
		\includegraphics[width=0.38\textwidth]{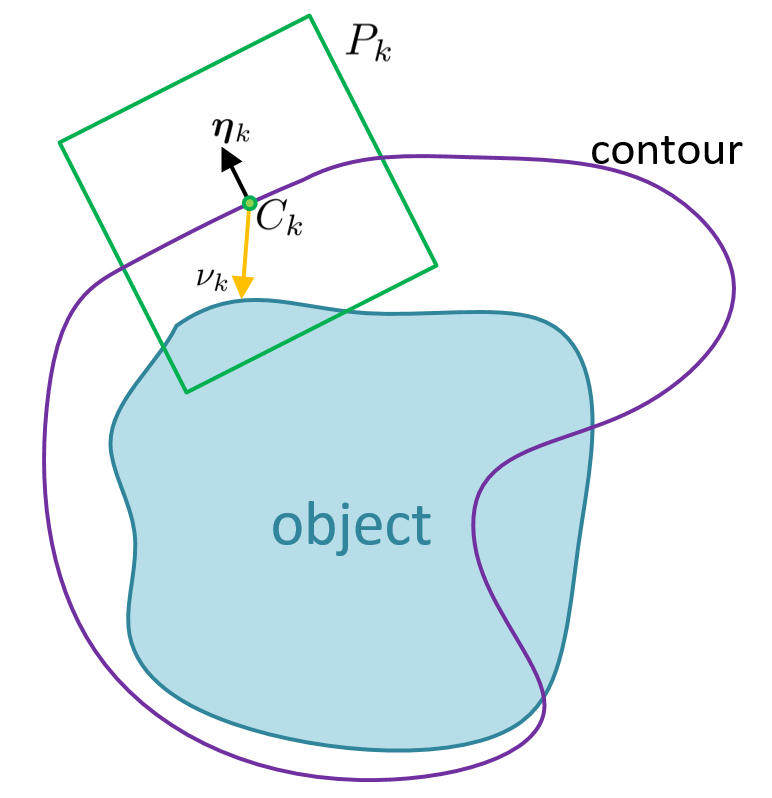}
	\end{center}
	\caption{\textbf{Patch extraction} Patches are extracted along the contour and oriented with respect to the normal.}
	\label{fig:patchContour}
\end{figure*}

Contrary to traditional active contour methods, where the contour is advected with a flow field $\nabla E(\C(s,t))$ in Equation \ref{eqn:gradFlow}, we learn to predict a flow field $\flow(s,t)$ using a CNN which is used for the evolution instead.
For each point $C_k$ on the contour, where $k=1,\dots,K$, we extract a small ($64 \times 64$) patch $\Patch_k$ such that the local coordinate system of the patch is aligned with the outer normal $\n_k$ at the respective position.
This way, we achieve rotation invariance since the orientation of a patch is now independent of the orientation of the object of interest.
Figure Fig.~\ref{fig:patchContour} shows the sampling of a patch $\Patch_k$ from a contour point $C_k$ w.r.t.~the normal $\n_k$ at that point.
These patches are then fed to a CNN which predicts a vector $\flow_k$ that points towards the closest object instance of the respective object category.
More precisely, let us define a mapping $f$ which maps a patch $\Patch_k$ to $\flow_k$, i.e. $f(\Patch) = \flow$.
This mapping will be learned using a CNN that will be described in detail in Sec.~\ref{sec:Architecture}.

For generating training data, we have to extract input-output pairs from a dataset for which we utilize ground truth segmentations.
Since the patch size is fixed to $64\times64$ pixels, we need to ensure that the extracted training patches capture the object.
Thus we turn the binary segmentation into a signed distance map (SDM) and consider all level lines in the range $[-15, 15]$.
Next, we iterative over all level lines and select a certain portion of contour points (per level line) for which patches are sampled.
The portion of selected contour points depends on the length $L$ of the respective level line and we use about $L/32$ contour points, where $L$ is measured in pixels.
In case data augmentation is needed, it has proven to be a good strategy to randomly rotate the extracted patches a randomly chosen angle in the range $\pm\pi/4$ and add a color or intensity bias to the extracted patch.
In order to take scale changes into account, we also extracted patches at three different scales: 100\%, 75\%, and 50\% of the original image size.
Besides sampling the patch, we also extract the gradient of the SDM at the respective point on the level line and scale it with the value of the SDM at this location.
This way, we obtain training pairs consisting of an image patch and a ground truth value for $\flow$.
At this point we wish to note the following points:
\begin{enumerate}
	\item It is important that the vector $\flow$ is extracted w.r.t. the local patch coordinate system, as shown in Fig. \ref{fig:patchContour}, in order to ensure rotation invariance.
	\item Although the patch size $64\times64$ appears to be quite small, it has proven to be a good trade-off between localization and context sensitivity. 
	As we will see in Section \ref{sec:Discussion}, the employed Sobolev-type regularization -- see \eqref{eqn:SobolevKernel} -- also benefits this compromise.
	\item While the runtime of other methods like \cite{long2015fully} scale linearly with the number of pixels, our method scales linearly only with the length $L$ of the curve, thus approximately logarithmically with the size of the object.
	As the convolution in \eqref{eqn:SobolevKernel} can be implemented in linear complexity (w.r.t. the number of contour points), the total complexity of our method is $\mathcal{O}(N\cdot L)$, where $N$ denotes the number of iterations as in Algorithm \ref{alg:DeepSnakeEvolution}.
\end{enumerate}
In the next section, we will now outline the network architecture.
%
\subsection{CNN Architecture}
\label{sec:Architecture}
\begin{figure*}[t]
	\centering
	\includegraphics[width=0.95\textwidth]{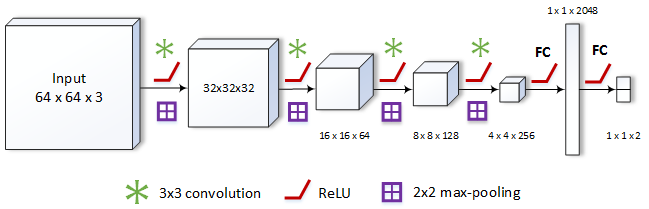}
	\caption{\textbf{CNN Architecture} The architecture of the CNN used to learn the mapping $f$ from input image $\I$ to the flow field $\flow$. In training, we minimize the standard L2 loss.}
	\label{fig:network}
\end{figure*}

Fig.~\ref{fig:network} shows the network architecture we used to model the mapping $f$.
It is inspired by the AlexNet architecture \cite{krizhevsky2012imagenet} but made simpler w.r.t.~size and structure for this regression task.
More precisely, it features an input size of $64 \times 64$ compared to originally $224 \times 224$.
It consists of four convolutional blocks, each with size $3 \times 3$, a padding of $1$ followed by a rectified linear unit and a $2 \times 2$ max-pooling step with stride 2.
Every convolutional block halves the feature map size but doubles the number of filters.
Thus the last feature map before the fully connected layer is of size $4 \times 4 \times 256$.
There is one fully connected layer with $2048$ neurons before the output layer of size $2$ which is the dimension of the predicted flow vector $\flow$.

The architecture can be chosen to be relatively simple as the amount of abstraction and discriminative power is expected to be much lower than in, for example, 1000 class image recognition.
\section{Experiments}
\label{sec:Experiments}
We evaluated the proposed method on two different data sets, i.e. the PASCAL VOC 2012 data set \cite{Everingham15} and the STACOM left ventricle segmentation data \cite{Suinesiaputra14}.
\subsection{Experimental Setup}
\label{sec:Setup}
On both datasets we split the data into training and testing sets and extract training patches as detailed in Section \ref{sec:Gradient}.
We train the proposed CNN architecture with a batch-size of 128 and a learning rate of $0.05$.
All networks are trained 20 epochs, choosing the one with lowest validation error, which is usually reached within 10 epochs.
Training converges in under two hours which is very short compared to other deep learning approaches.
\subsection{PASCAL VOC 2012}
\label{sec:Pascal}
As a first evaluation data set we chose the PASCAL visual object classification (VOC) database released in 2012, see Everingham \etal~\cite{Everingham15} for more information.
This data set includes images of objects from 20 different categories, i.e. aeroplane, bicycle, bird, boat, bottle, bus, car, cat, chair, cow, dining table, dog, horse, motorbike, person, potted plant, sheep, sofa, train and monitor/tv.
We decided to use this data, because it features a number of difficulties such as a large intra-class variability in terms of the depicted objects, no restrictions regarding the pose or viewpoint of the individual objects, cluttered backgrounds or various weather and lighting conditions.

We trained a network for each category as described in Section \ref{sec:Gradient} using the images of the testing data set.
Next, we traversed the validation data set in a similar manner and compared the predicted value of $\flow$ with the ground truth, i.e. the scaled gradient of the SDM.
This way, we are able to evaluate how well the proposed network, see Section \ref{sec:Architecture}, is able to predict the correct boundary location.
For this experiment, we evaluated more than nine million patches from images in the validation data set.
Tab.~\ref{tab:AngleStatistics} lists the percentage of predictions which do not divert from the ground truth by more then a fixed angle threshold.
We can see that in many categories more than two third of the predicted angles has an error of less than ten degrees.
Since the predicted flow $\flow$ is projected onto the normal $\n$, an error of less than 90 degrees still facilitates the evolution of the contour into the right direction.

\begin{figure*}[t]
	\begin{center}
		\includegraphics[width=0.38\textwidth]{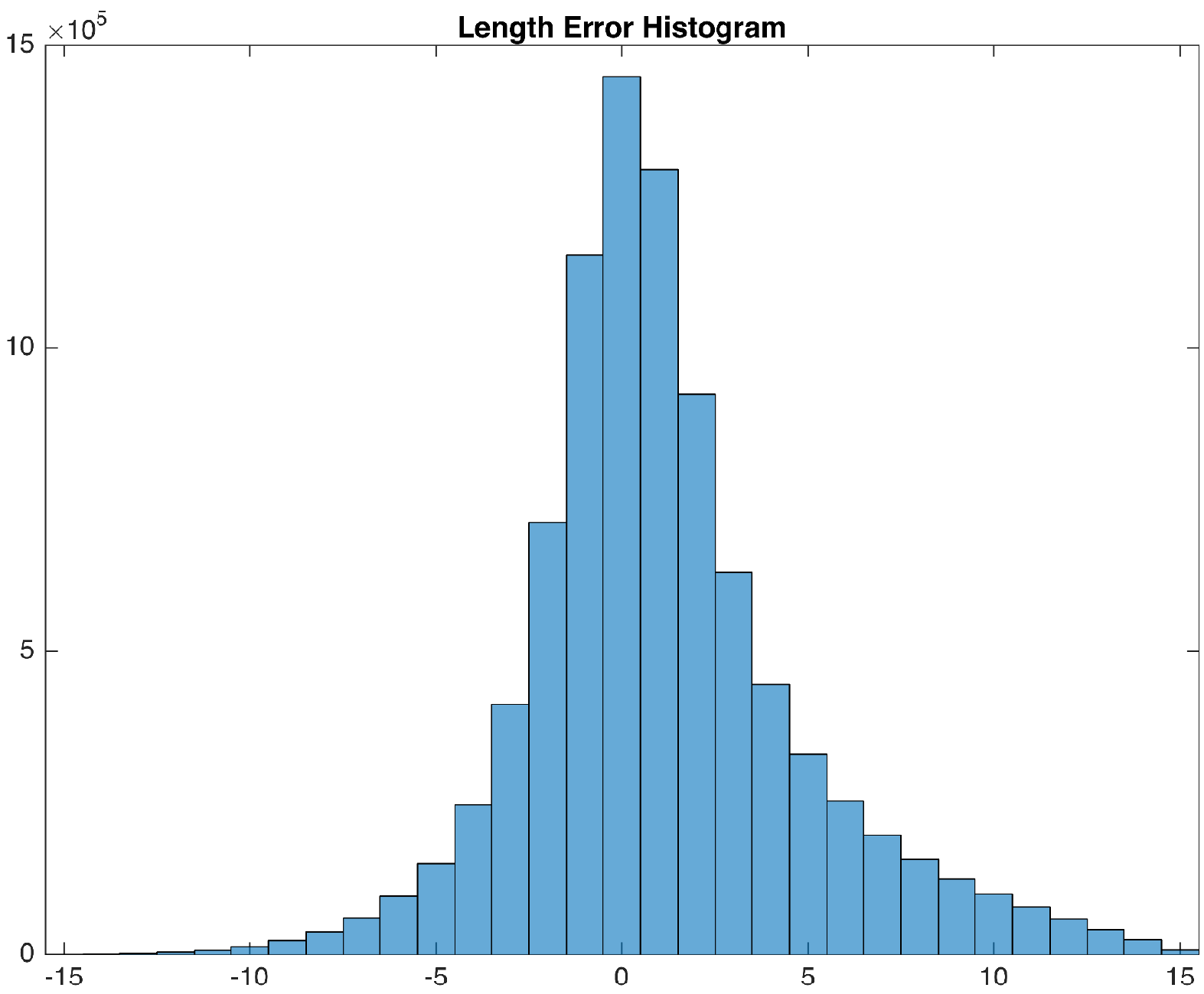}
		\caption{\textbf{Signed Length Error of Predictions} for all categories in the PASCAL VOC 2012 data set. The histogram reveals that the length of most of the vectors is predicted correctly or at least with a relatively small error.}
		\label{fig:LengthHistogram}
	\end{center}	
\end{figure*}

Fig.~\ref{fig:LengthHistogram} shows the signed error (in pixel) of the predicted length of $\flow$ in comparison to the ground truth.
One can observe that in most of the cases, the length of $\nu$ is predicted with an accuracy of less than 2 pixels.

It is important to note the following aspects:
\begin{enumerate}
\item Small angle deviations do not matter in practice, because we only use the projection of $\flow$ onto the normal $\n$ for evolving the contour, see Algorithm \ref{alg:DeepSnakeEvolution} in Section \ref{sec:Generalized}.
\item Even if a small portion of vectors is pointing into an entirely wrong direction or has the wrong length we did not observe any problems in practice, because the employed Sobolev-type regularization is effectively removing such outliers, see also the respective discussion in Section \ref{sec:Discussion}.
\end{enumerate}

\begin{figure*}[t]
	\includegraphics[height=0.24\textwidth]{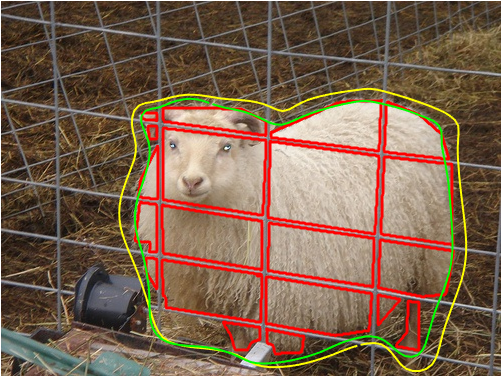}\hfill
	\includegraphics[height=0.24\textwidth]{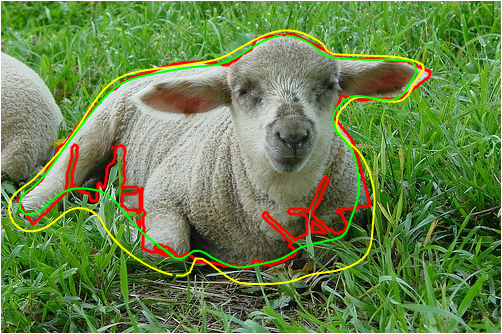}\hfill
	\includegraphics[height=0.24\textwidth]{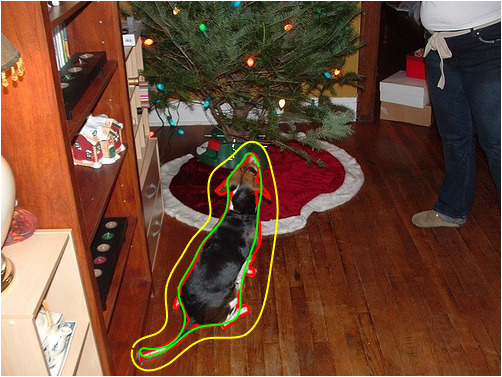}\\
	\includegraphics[height=0.2\textwidth]{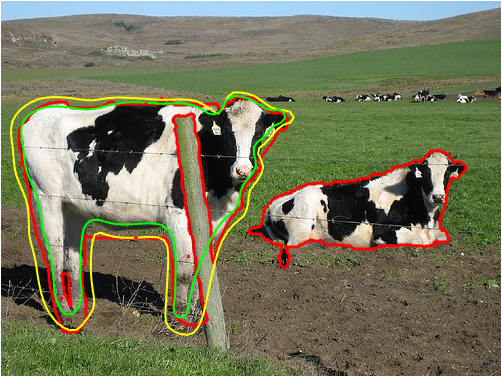}\hfill
	\includegraphics[height=0.2\textwidth]{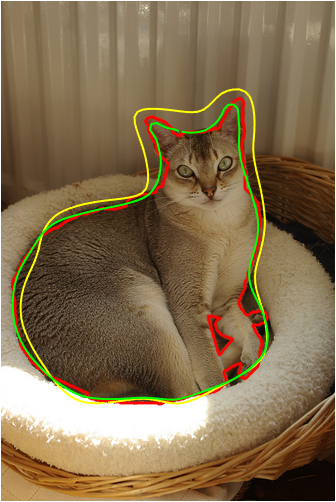}\hfill
	\includegraphics[height=0.2\textwidth]{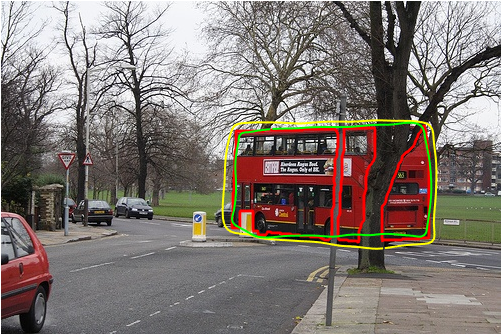}\hfill
	\includegraphics[height=0.2\textwidth]{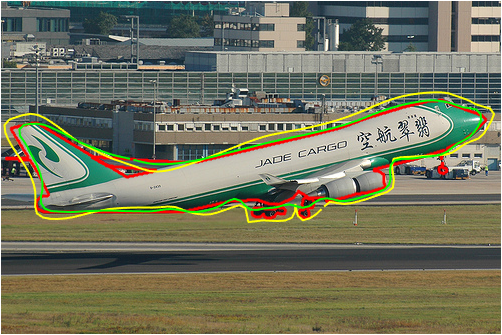}
	\caption{\textbf{Exemplary Results for the PASCAL VOC Dataset:} The proposed method work particularly well in case of occlusions (first sheep, cow, and bus) or additional instances of the same object class (second sheep and cow). The ground truth segmentation is shown in red, the contour initialization in yellow and our final segmentation in green.}
	\label{fig:PASCALexamples}
\end{figure*}

In addition to these experiments, we show a selection of segmentation results in Fig. \ref{fig:PASCALexamples}.
It can be observed that the proposed method performs well in case of fuzzy object boundaries, cluttered backgrounds, or multiple objected instances being present in one image.
\begin{table*}[t]
	\centering
	\caption{\textbf{Angle Error Statistics on PascalVOC} We compute the percentage of predicted flow vectors that differ only up to a specific threshold from the ground truth for all categories and overall}
	\begin{tabular}{c |*{21}r}
		\rot{Angle Thresh.} & \rot{\textbf{overall}} & \rot{aeroplane} & \rot{bicycle} & \rot{bird} & \rot{boat} & \rot{bottle} & \rot{bus} & \rot{car} & \rot{cat} & \rot{chair} & \rot{cow} & \rot{diningtable} & \rot{dog} & \rot{horse} & \rot{motorbike} & \rot{person} & \rot{potted plant} & \rot{sheep} & \rot{sofa} & \rot{train} & \rot{tv} \\ \midrule
		$<5^{\circ}$ &  49 &  52 &  44 &  51 &  46 &  52 &  53 &  47 &  53 &  48 &  49 &  47 &  50 &  50 &  45 &  50 &  41 &  49 &  52 &  47 &  54 \\ \midrule
		$<10^{\circ}$ &  68 &  71 &  64 &  71 &  63 &  70 &  70 &  65 &  73 &  67 &  68 &  63 &  70 &  70 &  63 &  70 &  58 &  69 &  70 &  65 &  69 \\ \midrule
		$<45^{\circ}$ &  84 &  87 &  86 &  86 &  81 &  82 &  81 &  80 &  88 &  83 &  86 &  73 &  86 &  85 &  80 &  87 &  79 &  88 &  84 &  79 &  79 \\ \midrule
		$<90^{\circ}$ &  85 &  88 &  88 &  87 &  82 &  82 &  81 &  81 &  88 &  84 &  87 &  74 &  87 &  86 &  81 &  88 &  81 &  89 &  86 &  80 &  81 \\ \midrule
	\end{tabular}
	\label{tab:AngleStatistics}
\end{table*}
\subsection{STACOM Challenge}
\label{sec:STACOM}
As a second application scenario, we consider the segmentation of the left ventricular cavity acquired with magnetic resonance imaging (MRI). 
For these experiments we used the data of the STACOM left ventricle challenge \cite{Suinesiaputra14}, which consists of 45 short axis MRIs acquired from 32 male and 13 female subjects.
We chose to evaluate the proposed method on this data set for a  number of reasons:
\begin{enumerate}
	\item Interactive segmentation methods are particularly interesting for medical applications, because the final segmentation results need to be approved by a medical expert in any case.
	\item The segmentation of the left ventricular cavity poses a lot of interesting challenges: Besides the fact that short axis MRIs of the heart are affected by flow, motion, off-resonance behavior and noise, the delineation of the boundary is particularly difficult in areas where the papillary muscles are present \cite{Suinesiaputra14}. 
	The latter difficulty is a very important one, because the papillary muscles have the same intensity as the myocardium.
	However, they actually need to be included in the blood pool characterized by bright intensities.
	\item Despite all these difficulties, the piecewise constant nature of the segmentation problem, i.e. the bright blood pool surrounded by dark myocardium enables us to employ a piecewise constant segmentation model, similar to the one considered by Chan and Vese \cite{Chan01}, as a baseline approach. 
\end{enumerate}
Our baseline approach consists of a standard Sobolev active contour as proposed by Sundaramoorthi \etal~\cite{Sun07} which uses the piecewise constant model
\begin{equation}
\int_{inside\,\C} (I(x) -\mu_i)^2\;dx+\int_{outside\,\C}(I(x)-\mu_o)^2\;dx,\quad \mu_i,\mu_o>0,
\end{equation}
as a data term. 
As such piecewise constant models are known to be susceptible to intensity variations and background clutter, we had to ensure that the comparison to the baseline approach is fair. 
Thus, we computed the mean value inside the contour $\mu_i$ from the ground truth segmentation of the left ventricular cavity and estimated the mean value outside the contour $\mu_0$ based on a narrow band of width 3 pixel around the ground truth segmentation.
We split the data into three sets consisting of 15 subjects and performed a threefold cross-validation. 
For each image, we took the ground truth segmentation and applied a random perturbation $\gamma(s)$ in normal direction which was computed according to
\begin{eqnarray}
\gamma(s) = \min(0.12\cdot \min(\Delta_x,\Delta_y),20))\cdot r_1\cdot \sin( 10\pi s r_2/L),
\end{eqnarray}
where $\Delta_x=\max_s(\C(s)_x)-\min_s(\C(s)_x)$ and $\Delta_y=\max_s(\C(s)_y)-\min_s(\C(s)_y)$ denote the extend of the curve in $x$ and $y$ direction and $r_1,r_2\in[0,1]$ are randomly generated numbers.
In case of a circular ground truth curve, the computed perturbation would be a sinusoidal perturbation in normal direction with frequency below five (with one period being the curve length $L$) and amplitude being below the minimum of 20 pixels and 12\% of the diameter.
For each image in the respective test set, we computed the result for ten different initializations, resulting in 8050 experiments.
Similar to \cite{Suinesiaputra14}, we compute the sensitivity $p$, the specificity $q$, the positive predictive value $PPV$, the negative predictive value $NPV$, and the Jaccard index $J$.
In addition to this, we compute the DICE score $D$.
The mean as well as the median of all these measures (w.r.t.~all 8050 experiments) is reported in Table \ref{tab:STACOM}.
When evaluating the results in Tab.~\ref{tab:STACOM}, it is very important to consider that the baseline uses the ground truth segmentation to estimate mean values of inside and outside of the object.
In a real application scenario this would not be possible, of course, but we wanted to compare to a baseline approach  both strong and conceptually comparable to the proposed method.
\begin{table*}[t]
	\centering
	\caption{\textbf{Quantitative Evaluation on the STACOM dataset:}We compare the dataset's standard error measures for our method to a baseline approach \cite{Sun07} which is allowed to access the ground truth mean for fore- and background to create a more fair, conceptually similar and strong baseline.}
	\begin{tabular}{lllrrrrrr}
		\toprule
		set& &method&$p$&$q$&$PPV$&$NPV$&$J$&$D$\\ \midrule
		set 1 $\quad$ & mean $\quad$  & baseline $\quad$ & $\quad$0.81 & $\quad$0.99 & $\quad$0.90& $\quad$1.00& $\quad$0.73& $\quad$0.83\\
		& & proposed & 0.85 & 0.99 & 0.93 & 1.00 & 0.79 & 0.86\\ \midrule
		set 1 & median & baseline & 0.84 &   1.00  &  0.98 &   1.00 &   0.80 &   0.89\\
		& & proposed & 0.91 &  1.00 &  0.99  & 1.00 & 0.86 & 0.93 \\ \midrule\midrule
		set 2 & mean & baseline & 0.81 & 0.99 & 0.90 & 1.00 & 0.73 & 0.83\\
		& & proposed & 0.85 & 0.99 & 0.93 & 1.00 & 0.79 & 0.86\\ \midrule 
		set 2 & median & baseline & 0.79 &    1.00 & 0.99 & 1.00 &  0.75 & 0.86\\
		& & proposed & 0.92 & 1.00 & 0.97 & 1.00 & 0.87 & 0.93 \\ \midrule\midrule
		set 3 & mean & baseline & 0.81 & 0.99 & 0.90 & 1.00 & 0.73 & 0.83\\
		& &  proposed & 0.85 & 0.99 & 0.93 & 1.00 & 0.79 & 0.86\\ \midrule
		set 3 & median & baseline & 0.83  &  1.00 &   0.98 &   1.00 &   0.80 &   0.89 \\
		& &  proposed & 0.85  &  1.00  &  1.00  &  1.00 &   0.82  &  0.90\\ \midrule\midrule
		%
		overall & mean & baseline & 0.80 & 0.99 & 0.89 & 1.00 & 0.72 & 0.81\\
		& & proposed & 0.84 & 0.99 & 0.92 & 1.00 & 0.78 & 0.85\\ \midrule
		overall & median & baseline & 0.82 & 1.00 & 0.98 & 1.00 & 0.78 & 0.88\\
		& & proposed & 0.90 & 1.00 & 0.99 & 1.00 & 0.85 & 0.92\\ \bottomrule
		
	\end{tabular}
	\vspace{0.2cm}
	\label{tab:STACOM}
\end{table*}
 \begin{figure*}[b]
 	\begin{center}
 		\includegraphics[width=0.45\textwidth]{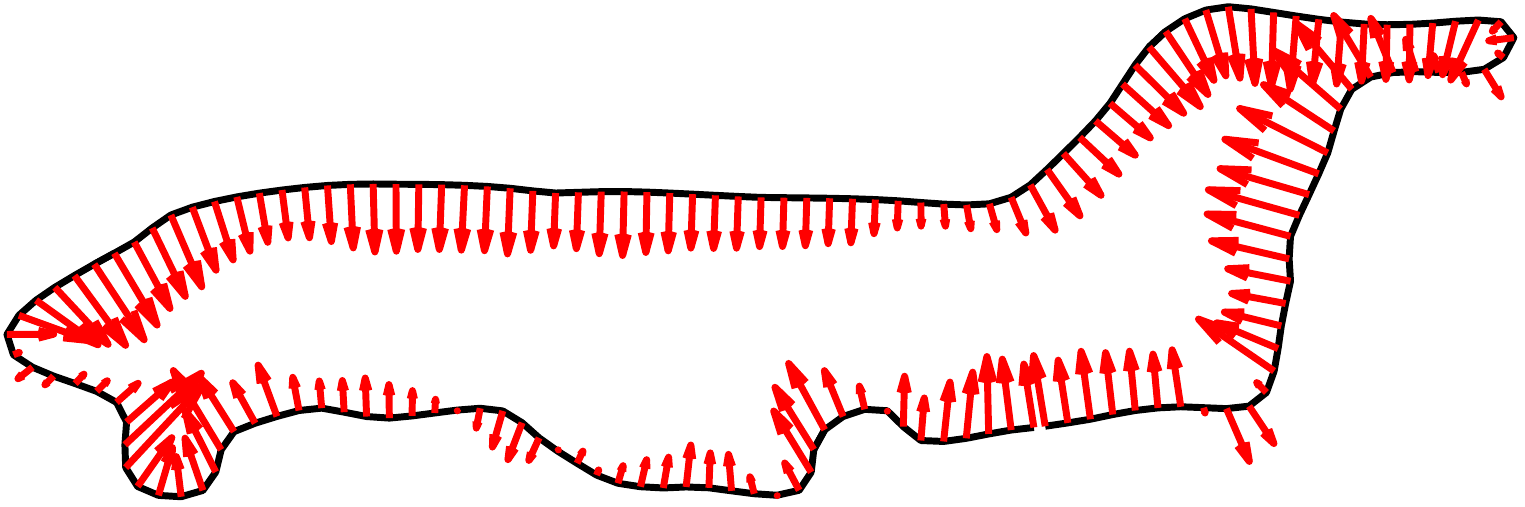}
 		\includegraphics[width=0.45\textwidth]{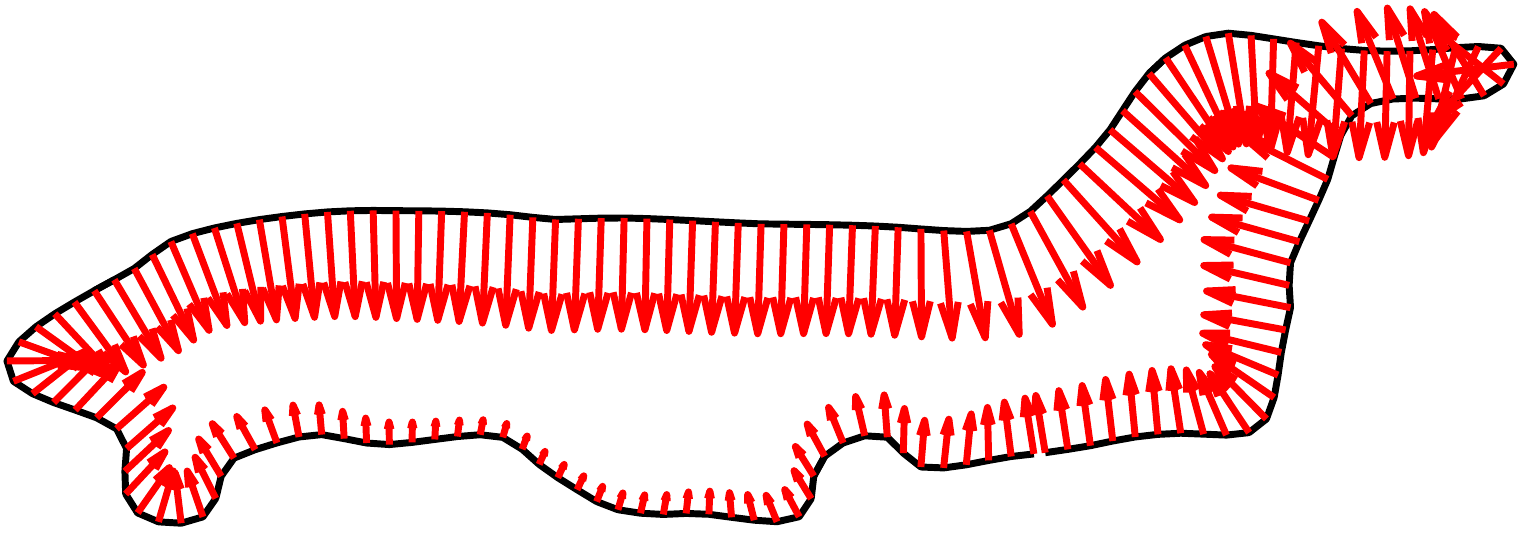}
 		\caption{\textbf{Effect of Sobolev-type Regularization:} The left panel shows the vector field $\flow$ as it is predicted by the CNN. The right panel shows the effect of the Sobolev-type regularization, cf. \eqref{eqn:SobolevKernel}. Please note how spurious predictions, e.g. at the lower right corner, are removed by this regularization.}
 		\label{fig:regularizingEffect}
 	\end{center}	
 \end{figure*}

\section{Discussion}
\label{sec:Discussion}
In this section we explain the beneficial effect of the employed regularization in connection to the experiments on the PASCAL VOC dataset in Section \ref{sec:RegularizingEffect}.
Furthermore, we wish to shed a light on what the network has learned by aggregating vote maps in Section \ref{sec:VoteMaps}.
Finally, we dicuss the limitations of the proposed approach in Section \ref{sec:Limitations}.
\subsection{Regularizing Effect}
\label{sec:RegularizingEffect}
As observed in Tab.~\ref{tab:AngleStatistics}, most of the predicted vectors have roughly the right direction.
However, there are still some vectors, especially the shorter ones, whose direction is completely wrong as depicted in the left panel of Fig. \ref{fig:regularizingEffect}.
By applying the proposed regularization in \eqref{eqn:SobolevKernel}, however, these spurious predictions can be effectively removed as shown in the right panel of Fig. \ref{fig:regularizingEffect}, which is one reason for the good performance of the proposed method.
\subsection{Aggregating Vote Maps}
\label{sec:VoteMaps}
\begin{figure*}[t]
	\begin{center}
		\includegraphics[width=0.98\textwidth]{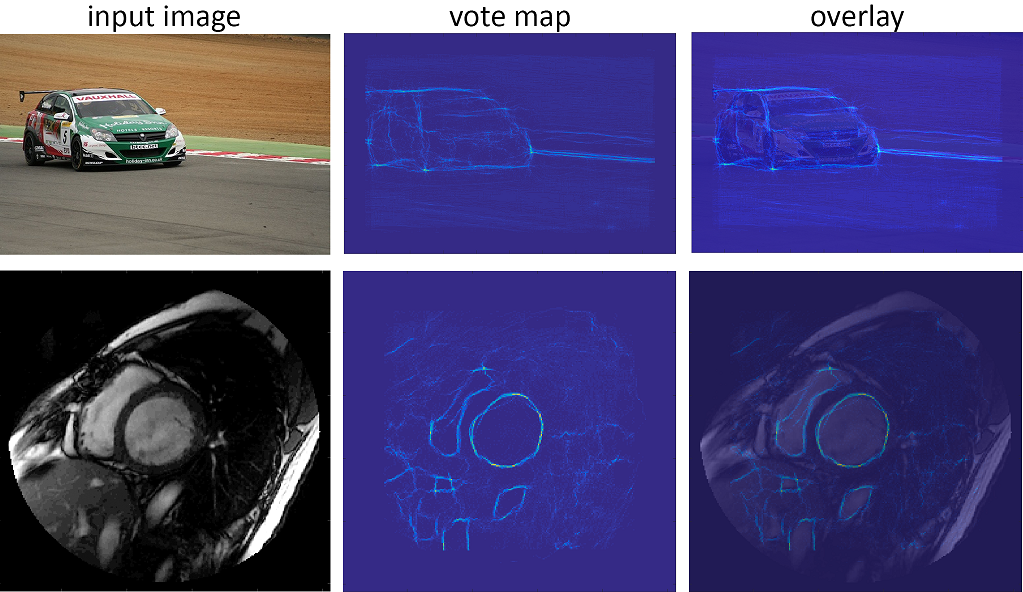}
		\caption{\textbf{Vote maps} the first row shows input image, vote map and an overlay of both on an image of the car category of PascalVOC. The second row contains an example from the STACOM dataset. Vote maps are obtained by predicting the flow in a dens sliding window and in four rotations across the whole image. The lighter the color the more votes for this location were collected.}
		\label{fig:votemaps}
	\end{center}	
\end{figure*}

To further asses the quality of our flow field, we predict a dense vector field for the whole image by evaluating the network at every pixel location, where we consider patches oriented w.r.t.~the four cardinal directions ($0^{\circ}$, $90^{\circ}$, $180^{\circ}$ and $270^{\circ}$).
As the CNN was trained to predict the offset vector pointing from the patch center towards the boundary of the closest object instance (for which the network has been trained), the network is essentially voting for the location of the closest object boundary. 
The rotations are necessary to account for the patches being aligned to the contour's outer normal. 
Thus, we can collect these votes and visualize them in a vote map, see Fig.~\ref{fig:votemaps}.
We observe the following aspects:
\begin{enumerate}
	\item As corner points on object corners have many pixels for which they are the closest ones, they accumulate the most votes.
	Since we project the predicted vectors onto the normal of the contour (see \eqref{eqn:speedFunction}) this does, however, not bias the evolution towards these points.
	\item It is interesting to observe that on the object itself, the actual three-dimensional object edges are voted for -- and not only the silhouette of the object.
	For an example, consider the left boundary of the windshield of the car in Fig.\ref{fig:votemaps}.
	This indicates that the network has learned, to a certain extend, the shape and transformations of the car by only learning from two-dimensional patches.
	\item In case of the medical example, it can be observed that the boundary of right ventricular cavity only accumulates many votes when it comes close to the boundary of the left ventricular cavity, which is the object of interest in this case.
	Thus, we can conclude that the proposed approach is capable of learning enough context information.	
\end{enumerate}
\subsection{Limitations}
\label{sec:Limitations}
The proposed methods has, of course, some limitations.
As it can be observed in Fig. \ref{fig:PASCALexamples}, that the method has problems when the boundary of the object has some details which cannot be accurately resolved by the comparatively coarse patches.
Consider for an example, the legs of the cow or the ears of the dog in Fig. \ref{fig:PASCALexamples}.
For this reason, future work will include investigations on how to solve this problem.

\section{Conclusion}
\label{sec:Conclusion}
We presented an interactive segmentation method based on the concept of active contours.
It employs a CNN predicting a vector field attached to the curve which is then used to evolve the contour towards the object boundary.
Furthermore, it relies on a Sobolev-type regularization, cf. \cite{Sun07}, for removing spurious predictions.

The architecture of the CNN is inspired by the network of Krizhevsky \etal \cite{krizhevsky2012imagenet}, but it is considerably smaller and thus very efficient from a computational point of view.
For an example, it can be trained on comparatively small GPUs such as a NVIDIA GTX 980 with only 4GB of RAM in less than two hours.

We evaluated the proposed method on the STACOM data set \cite{Suinesiaputra14} and the PASCAL VOC 2012 data set \cite{Everingham15}.
We were able to show the broad applicability of the proposed method as well as its potential for challenging applications such as the segmentation of the left ventricular cavity from short axis MRI scans.

Future work includes the evaluation of multi-class networks as well as the evaluation of coarse-to-fine approaches for increasing the accuracy.

\bibliographystyle{splncs}
\bibliography{references}
\end{document}